# Template Matching in Images using Segmented Normalized Cross-Correlation

Davor Marušić[1], Siniša Popović[2], Zoran Kalafatić[2]

[1]Independent Researcher, Zagreb, Croatia
[2]University of Zagreb Faculty of Electrical Engineering and Computing, Zagreb, Croatia
sinisa.popovic@fer.unizg.hr

**Abstract**: In this paper, a new variant of an algorithm for normalized cross-correlation (NCC) is proposed in the context of template matching in images. The proposed algorithm is based on the precomputation of a template image approximation, enabling more efficient calculation of approximate NCC with the source image than using the original template for exact NCC calculation. The approximate template is precomputed from the template image by a split-and-merge approach, resulting in a decomposition to axis-aligned rectangular segments, whose sizes depend on per-segment pixel intensity variance. In the approximate template, each segment is assigned the mean grayscale value of the corresponding pixels from the original template. The proposed algorithm achieves superior computational performance with negligible NCC approximation errors compared to the well-known Fast Fourier Transform (FFT)-based NCC algorithm, when applied on less visually complex and/or smaller template images. In other cases, the proposed algorithm can maintain either computational performance or NCC approximation error within the range of the FFT-based algorithm, but not both.

# 1. Introduction

Template matching is a technique often used in digital image processing, where it solves a problem of finding a smaller part of an image by comparing the image with a given template that contains a section or pattern that is searched for. Template matching involves the process of "sliding" the template over the image, meaning the template is moved $u$ discrete steps in the $x$ direction and $v$ steps in the $y$ direction of the image, and the calculation is performed over the template area for each position ($u,v$). Although AI-based methods have advanced the field of image processing, including pattern matching tasks (Buniatyan et al., 2017; Gao et al., 2024), for target patterns of fixed orientation and scale in tasks like automated testing of user interfaces, guiding bots in video games, or identifying fixed patterns in industrial quality control processes, satisfactory results can also be achieved by more traditional methods, such as those based on normalized cross-correlation (NCC).

Since NCC computation that cycles through all template pixels and corresponding image pixels is computationally expensive, the main goal of the proposed segmented NCC is to reduce the number of calculations at each ($u,v$) step, by using two ideas: (1) building a low-frequency template approximation, in line with the general concept of approximating the template function by expanding the zero-mean template to the weighted sum of rectangular basis functions (Briechle & Hanebeck, 2001); (2) efficient summing of pixel intensities in rectangular regions, based on the method of integral images (summed-area tables) from template matching and image processing literature (Lewis, 1995, 1995b).

A concrete variant of conceptual approach in (Briechle & Hanebeck, 2001) is proposed herein, which constructs a split-and-merge segmented approximation of the template, where each rectangular segment of the template has a standard deviation of its pixel intensities below a predefined threshold and consequently gets assigned their mean value. Representing the template image with such rectangular segments allows segment-by-segment computations, instead of pixel-by-pixel, with the potential to reduce the number of steps in NCC calculation considerably. The precision and speed of this algorithm largely depend on the setting of the threshold for the maximum allowed standard deviation of each segment, where a higher allowed deviation leads to a lower number of segments, increasing speed and lowering precision, while lower allowed deviation leads to a higher number of segments, decreasing speed and increasing precision.

Section 2 of this paper covers NCC and gives key points of a well-known Fast Fourier Transform (FFT)-based algorithm (Lewis, 1995, 1995b) for efficient NCC computation. Sections 3 and 4 explain the proposed segmented template image approximation for efficient computation of approximate NCC. In Section 5, the corresponding algorithm is presented in high-level pseudocode, which precomputes the template approximation and uses it in a two-step coarse-fine manner (Rosenfeld & Vanderbrug, 1977) to find template matches in the source image. Section 6 presents experimental results of the proposed algorithm and compares its performance to the FFT-based algorithm.

## 2. Normalized Cross-Correlation

An image of size $M \times N$ may be defined as a function $f$ over spatial coordinates $(x, y)$, $x \in \{0,...,M-1\}$, $y \in \{0,...,N-1\}$, where $f(x,y)$ at any pair of coordinates represents the pixel intensity or value at that point. Since grayscale images are used for image processing problems like these, a single grayscale value is represented by $f(x,y) \in \{0,...,255\}$. RGB images are represented as three values $r(x,y)$, $g(x,y)$, $b(x,y) \in \{0,...,255\}$, but using a single grayscale value is more computationally efficient, which is why grayscale-converted images are used.

The pattern to be matched is a given grayscale-converted template image $t$ of the size $W \times H$, where pixel intensities are represented as $t(x,y)$, $x \in \{0,...,W-1\}$, $y \in \{0,...,H-1\}$.

The common way to determine position $(u^*, v^*)$ of pattern $t$ in source image $f$ is to calculate the NCC value $\rho$ between $f$ and $t$ at each pixel point $(u,v)$, and find the position with the highest $\rho$. Basic definition for NCC is given by

$$\rho(u,v) = \frac{\sum_{x,y}(f(x,y)-\bar{f}_{u,v})(t(x-u,y-v)-\bar{t})}{\sqrt{\sum_{x,y}(f(x,y)-\bar{f}_{u,v})^2 \sum_{x,y}(t(x-u,y-v)-\bar{t})^2}}, \quad (1)$$

where $\bar{f}_{u,v}$ denotes the mean value of $f(x,y)$ within the area of the template size $W \times H$ shifted to position $(u,v)$ in the source image. If NCC is computed in a naïve way, the number of required calculations is proportional to $W \cdot H \cdot (M-W+1) \cdot (N-H+1)$.

The denominator in Eq. (1) can be efficiently calculated by the method presented in (Lewis, 1995, 1995b), using sum-tables containing integrals (running sums) of pixel intensities and squared pixel intensities of the source image. This reduces the denominator equation complexity by the factor of $W \cdot H$ minus the cost related to the computation of sum-tables. NCC numerator is interpreted as a convolution of the zero-mean source and reversed template image, which is then efficiently computed using FFT. The overall complexity of Lewis' approach is dominated by FFT and amounts to $O(M \cdot N \cdot \log(M \cdot N))$. In general, when either the template image or the source image beneath the template is uniform (i.e., all pixels have the same color value), the NCC numerator and denominator both become zero, making NCC-based template matching unsuitable for such cases. Therefore, handling these cases is beyond the scope of either the FFT-based or the proposed algorithm in this paper.

While the method presented in (Lewis, 1995, 1995b) is used as the reference comparison method in the results section of this paper, it should be noted that template matching literature abounds with approaches invented over the past decades (Hashemi et al., 2016). Despite the relatively long history of template matching, novel methods are being actively developed, founded in theoretical advancements (Fageot et al., 2021; Almira et al., 2024) and data-driven deep learning advancements (Buniatyan et al., 2017; Gao et al., 2024). Even methods that revolve around NCC computation are an active research topic over the last fifty years to the present day (Barnea & Silverman, 1972; Lewis, 1995, 1995b; Rosenfeld & Vanderbrug, 1977; Goshtasby et al., 1984; Yoshimura &Kanade, 1994; Briechle & Hanebeck, 2001; Tsai et al., 2002; Sun et al., 2003; Cho et al., 2010; Yang et al., 2024).

# 3. Segmented Approximation of the Template Image

Since the template image $t$ is provided before the template matching process, all the necessary calculations on $t$ that are independent of the source image $f$ should be precomputed to accelerate this process. In this regard, the calculation of NCC denominator according to Eq. (1) should precompute the sum related to the template. Similarly, efficient calculation of NCC numerator via the FFT-based algorithm includes transforming $t$ to frequency domain and calculating the conjugate of it, which eliminates the necessity of repeating these calculations at each $(u,v)$ step.

To calculate the NCC numerator according to the approach proposed in this paper, a few template preprocessing steps are needed. The first step includes splitting the template image into rectangular segments $S_1, \ldots, S_K$, each with dimensions $W_i \times H_i$ and standard deviation of pixel intensities ($\sigma_i$) below the proposed threshold $\sigma_{max}$. This constraint for standard deviation of each segment is given by

$$\sigma_i = \sqrt{\frac{\sum_{(x,y) \in S_i}(t(x,y) - \bar{t}_i)^2}{W_i H_i}} < \sigma_{max}, \quad (2)$$

where $\bar{t}_i$ denotes the arithmetic mean of template pixel values belonging to the segment $S_i$. Starting with the whole template image as the initial segment, any segment that does not satisfy this constraint is split into two and the process is recursively repeated on each new smaller segment. All pixels within each constraint-satisfying segment $S_i$ are assigned the same value $\bar{t}_i$. The $\sigma_{max}$ threshold should always be lower than the standard deviation $\sigma_t$ of the whole template, otherwise the template will be represented as a single segment of uniform pixel values.

For the purpose of this paper, a recursive binary splitting procedure has been used, where each segment is split in half on the larger of its two dimensions, width or height. In order to prevent the occurrence of neighboring segments that have equal pixel intensities, a procedure to merge such segments is executed after splitting has been completed. Further details of this split-and-merge algorithm are given in section 5.

As the result of this algorithm, template $t$ has been decomposed into segments $S_i$ that have each been assigned the mean value of the template pixel intensities contained within the boundaries of the respective segment. Accordingly, segmented approximation $k$ of the template image $t$ is defined as

$$k(x,y) = \begin{cases} k_1, (x,y) \in S_1 \\ \vdots \\ k_K, (x,y) \in S_K \end{cases} \quad (3)$$

where $k_i$ is calculated by

$$k_i = \frac{1}{|S_i|} \sum_{(x,y) \in S_i} t(x,y) = \frac{1}{W_i H_i} \sum_{(x,y) \in S_i} t(x,y). \quad (4)$$

Summing all pixel intensities of the segmented template approximation can be done efficiently segment by segment:

$$\sum_{x,y} k(x,y) = \sum_{i=1}^{K} \left( \sum_{(x,y) \in S_i} k_i \right) = \sum_{i=1}^{K} W_i H_i k_i. \quad (5)$$

For computational efficiency, the approximate template is not stored as an image (a grid of pixel values), but rather as a vector of 5-tuples that represent starting coordinates and sizes of individual segments in the template image coordinate space and their uniform pixel values:

$$[S_1, S_2, \ldots, S_K] = [(x_1, y_1, W_1, H_1, k_1), (x_2, y_2, W_2, H_2, k_2)\ldots(x_K, y_K, W_K, H_K, k_K)]. \quad (6)$$

## 4. Efficient Calculation of Approximate Normalized Cross-Correlation

Replacing the template image $t$ with its approximation $k$ in Eq. (1) for the calculation of the NCC implies replacing the term $t(x - u, y - v) - \bar{t}$ with $k(x - u, y - v) - \bar{k}$, where $\bar{k}$ is calculated by

$$\bar{k} = \frac{\sum_{x,y} k(x,y)}{W \cdot H} = \frac{\sum_{i=1}^{K} W_i H_i k_i}{W \cdot H}. \quad (7)$$

Accordingly, calculation of the NCC numerator in Eq. (1),

$$num(u, v; f, t) = \sum_{x,y} (f(x,y) - \bar{f}_{u,v})(t(x - u, y - v) - \bar{t}), \quad (8)$$

can be done more efficiently by substituting the template $t$ with the segmented template approximation $k$, and partitioning the source image $f$ underneath the template based on segments $S_i$, in the following way:

$$num(u, v; f, t) \approx num(u, v; f, k) = \sum_{i=1}^{K} \left( \left( \sum_{(x,y) \in S_i} f(x,y) \right) - \bar{f}_{u,v} W_i H_i \right) (k_i - \bar{k}). \quad (9)$$

To optimize the calculation of $\sum_{(x,y) \in S_i} f(x,y)$ in Eq. (9), the sum-table (integral image) $s(u,v)$ over source image $f$ is used, which is defined by (Lewis, 1995, 1995b):

$$s(u, v) = f(u, v) + s(u - 1, v) + s(u, v - 1) - s(u - 1, v - 1). \quad (10)$$

Thus, $s(u, v)$ represents the sum of the whole rectangular area of the source image intensities $f(x, y)$ that are above and left of $(u, v)$, including $f(u, v)$ itself.

With the sum-table prepared, $\sum_{(x,y) \in S_i} f(x,y)$ can be efficiently calculated using similar algorithm:

$$\sum_{(x,y) \in S_i} f(x,y) = s(u + x_i - 1, v + y_i - 1) - s(u + x_i - 1, v + y_i + H_i - 1) - s(u + x_i + W_i - 1, v + y_i - 1) + s(u + x_i + W_i - 1, v + y_i + H_i - 1). \quad (11)$$

After using sum tables to calculate $\sum_{(x,y) \in S_i} f(x,y)$, we get

$$num(u, v; f, k) = \sum_{i=1}^{K} \bigl( s(u + x_i - 1, v + y_i - 1) - s(u + x_i - 1, v + y_i + H_i - 1) - s(u + x_i + W_i - 1, v + y_i - 1) + s(u + x_i + W_i - 1, v + y_i + H_i - 1) - \bar{f}_{u,v} W_i H_i \bigr) (k_i - \bar{k}), \quad (12)$$

which is the final calculation of the numerator from Eq. (1). The sum-table is used to efficiently compute the sum of pixel intensities needed to calculate $\bar{f}_{u,v}$:

$$\sum_{x,y} f(x,y) = s(u + W - 1, v + H - 1) - s(u - 1, v + H - 1) - s(u + W - 1, v - 1) + s(u - 1, v - 1). \quad (13)$$

For the calculation of the denominator from Eq. (1), we use the same $s(u, v)$ sum-table represented with Eq. (10) that has been used in numerator calculation, and additionally use $s^2(u, v)$ squared values sum-table over the source image $f$, given by (Lewis, 1995, 1995b):

$$s^2(u,v) = f^2(u,v) + s^2(u-1,v) + s^2(u,v-1) - s^2(u-1,v-1). \quad (14)$$

Similar to Eq. (13) used in calculation of the NCC numerator, with just 2 subtractions and 1 addition, sum of squared values $f^2(x,y)$ under the template image shifted at position $(u, v)$ of the source image can be calculated:

$$\sum_{x,y} f^2(x,y) = s^2(u + W - 1, v + H - 1) - s^2(u - 1, v + H - 1) - s^2(u + W - 1, v - 1) + s^2(u - 1, v - 1). \quad (15)$$

Eq. (13) and Eq. (15) are then used to compute the part of the NCC denominator related to source image $f$, because it can be rewritten as

$$\sum_{x,y} (f(x,y) - \bar{f}_{u,v})^2 = \sum_{x,y} f^2(x,y) - 2\bar{f}_{u,v} \sum_{x,y} f(x,y) + \sum_{x,y} \bar{f}_{u,v}^2 =$$

$$\sum_{x,y} f^2(x,y) - WH \bar{f}_{u,v}^2 = \sum_{x,y} f^2(x,y) - \bar{f}_{u,v} \cdot \sum_{x,y} f(x,y). \quad (16)$$

Eq. (16) reuses $\bar{f}_{u,v}$ and $\sum_{x,y} f(x,y)$, both of which have already been computed efficiently based on Eq. (13).

Accordingly, the denominator of NCC from Eq. (1) is approximated by

$$denom(u,v; f,k) = \sqrt{\left(\sum_{x,y} f^2(x,y) - \bar{f}_{u,v} \cdot \sum_{x,y} f(x,y)\right) \cdot \sum_{i=1}^{K} (k_i - \bar{k})^2}, \quad (17)$$

where the last sum is precomputed from the zero-mean variant of template approximation $k$.

Finally, the calculation of NCC between the source image $f$ and template image $t$ via substitution of the segmented template approximation $k$ relies on Eq. (12) and Eq. (17):

$$\rho(u,v) = \rho(u,v; f,t) \approx \rho(u,v; f,k) = num(u,v; f,k)/denom(u,v; f,k). \quad (18)$$

# 5. Proposed Template Matching Algorithm

The algorithm for efficient template matching based on NCC computation between the source image $f$ and the template image $t$ precomputes the segmented template approximation $k$ and computes $\rho(u, v; f, k)$ in a two-stage coarse-fine manner for all eligible $(u, v)$ within the source image coordinate space, according to Eq. (18). The algorithm proceeds as follows:

- Inputs: source image $f$ of size $M \times N$; template image $t$ of size $W \times H$
- Experimental thresholds: $\sigma_{max,fast} = 0.99\ \sigma_t$; $\sigma_{max,slow} = 0.1\ \sigma_t$; $K_{max} = 5000$; $precision = 0.9$
- Output: a collection *matches* of $(u, v, \rho)$ triplets denoting all matches of template $t$ in source image $f$ and the corresponding values of $\rho(u, v; f, k)$, initially empty
1. $(k_{fast}, \rho_{fast})$ := precompute_template_approximation($t$, $\sigma_{max,fast}$, $K_{max}$): // coarse approximation
2. $(k_{slow}, \rho_{slow})$ := precompute_template_approximation($t$, $\sigma_{max,slow}$, $K_{max}$): // fine approximation
3. $(k_{fast0}, k_{slow0})$ := precompute zero mean variants of $k_{fast}$ and $k_{slow}$
4. precompute $\sum_{i=1}^{K_{fast0}}(k_{fast0,i})^2$ and $\sum_{i=1}^{K_{slow0}}(k_{slow0,i})^2$
5. $threshold_{fast}$ := $precision \cdot \rho_{fast}$; $threshold_{slow}$ := $precision \cdot \rho_{slow}$
6. $(s, s^2)$ := compute_sum_tables($f$) using Eq. (10) and Eq. (14)
7. for $(u, v)$ in $\{0, …, M−W\} \times \{0, …, N−H\}$

    a) compute $\Sigma f$ and $\Sigma f^2$ for template position $(u, v)$ using Eq. (13) and Eq. (15)
    b) $\overline{f}_{u,v} := \Sigma f / (W \cdot H)$
    c) $num$ := $num(u, v; f, k_{fast})$ according to Eq. (12)
    d) $denom$ := $denom(u, v; f, k_{fast})$ according to Eq. (17), using precomputed $\sum_{i=1}^{K_{fast0}}(k_{fast0,i})^2$
    e) $\rho$ := $num/denom$
    f) if $(\rho < threshold_{fast})$ continue // too low NCCs with coarse template approximation are not candidates for a match
    g) $num$ := $num(u, v; f, k_{slow})$ according to Eq. (12)
    h) $denom$ := $denom(u, v; f, k_{slow})$ according to Eq.(17), using precomputed $\sum_{i=1}^{K_{slow0}}(k_{slow0,i})^2$
    i) $\rho$ := $num/denom$
    j) if $(\rho \geqslant threshold_{slow})$ append *matches* with $(u, v, \rho)$

Performance analyses of this algorithm versus FFT-based algorithm for NCC computation have been done by timing the execution of lines 1-5 as template preparation phase and lines 6, 7a-j as actual template matching phase.

To complete the algorithm, the remaining text in this section contains the pseudocode of the function precompute_template_approximation, and its key worker functions.

function precompute_template_approximation
- Inputs: template image $t$; threshold $\sigma_{max}$ controlling segment splitting when building the segmented template approximation; the maximum allowed number of segments $K_{max}$
- Outputs: $(k, \rho_{max})$, i.e. the segmented template approximation and its corresponding NCC with the template image $t$
1. repeat
    a) create empty vector $k$
    b) split_segment($t$, $\sigma_{max}$, $(0, 0, W, H)$, $k$)
    c) $\sigma_{max} := \sigma_{max} + 1$
2. until $(length(k) \leq K_{max})$
3. merge_redundant_segments($k$)
4. compute $\overline{k}, \overline{t}$
5. $num := \sum_{i=1}^{K}(\sum_{(x,y)\ \epsilon\ S_i} t(x,y) - \overline{t}W_i H_i))\ (k_i - \overline{k})$// equivalent of $\sum_{i=1}^{K}(\sum_{(x,y)\ \epsilon\ S_i}(t(x,y) - \overline{t}))\ (k_i - \overline{k})$

6. $denom := \sqrt{\sum_{x,y}(t(x,y) - \overline{t})^2 \sum_{i=1}^{K}(k_i - \overline{k})^2}$
7. return ($k$, $num/denom$)

function split_segment
- Inputs: template image $t$; threshold $\sigma_{max}$; segment to split given as ($x$, $y$, $width$, $height$); segmented template approximation $k$
- Outputs: updated $k$
1. if ($width$ = 1 and $height$ = 1) append ($x$, $y$, $width$, $height$, $t(x,y)$) to $k$; return
2. compute mean $\mu$ and standard deviation $std$ within this segment
3. if ($std < \sigma_{max}$) append ($x$, $y$, $width$, $height$, $\mu$) to $k$; return
4. if ($width > height$)
    a) split_segment($t$, $\sigma_{max}$, ($x$, $y$, $\lceil width/2 \rceil$, $height$), $k$)
    b) split_segment($t$, $\sigma_{max}$, ($x+ \lceil width/2 \rceil$, $y$, $\lfloor width/2 \rfloor$, $height$), $k$)
5. else
    a) split_segment($t$, $\sigma_{max}$, ($x$, $y$, $width$, $\lceil height/2 \rceil$), $k$)
    b) split_segment($t$, $\sigma_{max}$, ($x$, $y+\lceil height/2 \rceil$, $width$, $\lfloor height/2 \rfloor$), $k$)

function merge_redundant_segments
- Inputs: segmented template approximation $k$
- Outputs: updated $k$
1. sort $k = [(x, y, w, h, \mu)_i]$ in the ascending lexicographic order on the first and second dimension, i.e. on $x$ and $y$ coordinates of segments' top-left corners
2. merge segments along $y$ axis (vertically):
    a) $i := 0$
    b) while ($i <$ length($k$)−1)
        i. if ($k[i].w = 0$ or $k[i].h = 0$) $i := i+1$; continue // skip any degenerate segments
        ii. $j := i+1$
        iii. while ($j <$ length($k$) and $k[i].x = k[j].x$ and $k[i].w = k[j].w$ and $k[i].y+k[i].h = k[j].y$ and $k[i].\mu = k[j].\mu$)
            1. $k[i].h := k[i].h + k[j].h$
            2. $k[j].h := 0$
            3. $j := j+1$
        iv. $i := j$
3. sort $k = [(x, y, w, h, \mu)_i]$ in the ascending lexicographic order on the second and first dimension, i.e. on $y$ and $x$ coordinates of segments' top-left corners
4. merge segments along $x$ axis (horizontally), using code block analogous to 2a-b
    a) $i := 0$
    b) while ($i <$ length($k$)−1)
        i. if ($k[i].w = 0$ or $k[i].h = 0$) $i := i+1$; continue // skip any degenerate segments
        ii. $j := i+1$
        iii. while ($j <$ length($k$) and $k[i].y = k[j].y$ and $k[i].h = k[j].h$ and $k[i].x+k[i].w = k[j].x$ and $k[i].\mu = k[j].\mu$)
            1. $k[i].w := k[i].w + k[j].w$
            2. $k[j].w := 0$
            3. $j := j+1$
        iv. $i := j$
5. remove segments with width or height equal to 0
6. if $k$ has been changed in step 5, go to step 1; else return

# 6. Comparative Results of the Proposed and FFT-based Template Matching Algorithms

The computational efficiency of the template matching algorithm proposed in this paper, using two segmented template approximations with high and low allowed standard deviation of pixel intensities per segment (fast/coarse and slow/fine approximation) is influenced by a number of variables. Firstly, it depends on image size $M \times N$ and on number of segments $K$. The algorithm works in a way that search is made with fast approximation of the template, with $K_{fast}$ number of segments, while search with slow approximation, having $K_{slow}$ number of segments, is executed only when the NCC with fast approximation is above a predefined threshold. If $Q$ denotes the number of times slow search is executed, complexity of the algorithm is $K_{fast} \cdot (M-W+1) \cdot (N-H+1) + Q \cdot K_{slow}$, and the worst-case scenario is $(K_{fast}+K_{slow}) \cdot (M-W+1) \cdot (N-H+1)$. For template images with very high visual complexity (Yu & Winkler, 2013; Cardaci et al., 2009), like random noise images, the number of segments may approach the number of pixels, so the maximum allowed number of segments $K_{max}$ was proposed to sacrifice algorithm precision for speed. However, at certain template size and the level of visual complexity switching to FFT-based approach should be more computationally efficient.

The proposed algorithm and FFT-based algorithm were comparatively tested using identical source images and templates on a 3.6GHz 8-core AMD Ryzen 7 3700X processor. Algorithms were developed in Rust programming language utilizing parallelism (Rayon library), while FFT-based algorithm additionally used RustFFT library. The source code with evaluation results is available at: https://github.com/DavorMar/rustautogui. Images used in the comparative analysis of these template matching algorithms are shown in the Appendix (Figure 1, Figure 2 and Figure 3), as to the best of our knowledge there is no open dataset of semantically diverse images designed specifically for benchmarking template matching algorithms. Table 1 shows results of an evaluation of the two template matching algorithms, and a naïve template search algorithm using exact pixelwise NCC computation according to Eq. (1). Table 2 shows template preparation times related to the construction of segmented template approximation for the proposed algorithm, and to conjugate FFT of the template zero-padded to the source-image next power of two for the FFT-based algorithm.

While the proposed algorithm performs comparably well on Socket (Figure 1) and Darts (Figure 2) images, Split city source image (Figure 3) has been included to present limitations of the proposed algorithm when searching for a large and visually complex template image, like templates t2 and t3 (underlined results in Table 1). Template t5 illustrates that the proposed algorithm can perform well in comparison to FFT-based algorithm on large template images which have lower visual complexity. Naïve search testing on Split city templates t3 and t5 has been stopped because the process took more than 40 minutes, as these exact timing results are irrelevant. Finally, when comparing results of the proposed algorithm to OpenCV configured with single-instruction multiple-data CPU extensions but without OpenCL support, template search using the proposed algorithm was 1.1–7.0 times slower, except for Split city template t5 where OpenCV template search was 1.2 times slower.

Table 1: Evaluation results of the proposed and FFT-based template matching algorithms

| Source image ($M \times N$) | Template image ($W \times H$) | Segments count in $k_{fast}, k_{slow}$ | Search duration (proposed algorithm) [s] | Search duration (FFT-based algorithm) [s] | Search duration (naïve algorithm) [s] | Max. computed NCC (proposed algorithm) | Max. computed NCC (FFT-based algorithm) |
|---|---|---|---|---|---|---|---|
| Socket (539×959) | template 1 (154×143) | 119, 4717 | **0.034** | 0.084 | 1.660 | 0.99 | 1.0 |
| | template 2 (110×73) | 118, 1474 | **0.038** | 0.089 | 0.725 | 0.99 | 1.0 |
| | template 3 (130×135) | 27, 1871 | **0.023** | 0.091 | 1.433 | 0.99 | 1.0 |
| Darts (538×508) | template 1 (159×156) | 32, 4790 | **0.009** | 0.084 | 0.780 | 0.91 | 1.0 |
| | template 2 (376×226) | 121, 4771 | **0.008** | 0.078 | 0.914 | 0.8 | 1.0 |
| | template 3 (83×31) | 11, 741 | **0.008** | 0.081 | 0.139 | 0.98 | 1.0 |
| Split city (4032×3024) | template 1 (82×77) | 89, 987 | **0.799** | 1.518 | 17.690 | 0.99 | 1.0 |
| | template 2 (1162×680) | 5, 4420 | <u>**1.951**</u> | <u>1.457</u> | 1223.059 | 0.75 | 1.0 |
| | template 3 (2990×1938) | 3, 3973 | <u>**2.020**</u> | <u>1.518</u> | > 40 min | 0.74 | 1.0 |
| | template 4 (80×118) | 41, 2706 | **1.094** | 1.607 | 25.689 | 0.95 | 1.0 |
| | template 5 (3908×860) | 12, 1222 | **0.228** | 1.483 | > 40 min | 0.98 | 0.968 |

Table 2: Template preparation times for the proposed and FFT-based template matching algorithms

| Template image ($W \times H$) | Proposed algorithm [s] | FFT-based algorithm[s] |
|---|---|---|
| Socket template 1 (154×143) | 0.012 | 0.036 |
| Socket template 2 (110×73) | 0.002 | 0.037 |
| Socket template 3 (130×135) | 0.006 | 0.040 |
| Darts template 1 (159×156) | 0.104 | 0.040 |
| Darts template 2 (376×226) | 0.334 | 0.036 |
| Darts template 3 (83×31) | 0.006 | 0.039 |
| Split city template 1 (82×77) | 0.002 | 0.621 |
| Split city template 2 (1162×680) | 1.450 | 0.639 |
| Split city template 3 (2990×1938) | 10.541 | 0.671 |
| Split city template 4 (80×118) | 0.017 | 0.655 |
| Split city template 5 (3908×860) | 0.523 | 0.662 |

# Appendix

Appendix contains source images and templates with their 2 segmented approximations related to the proposed algorithm.

**Source image**    t1)    t2)

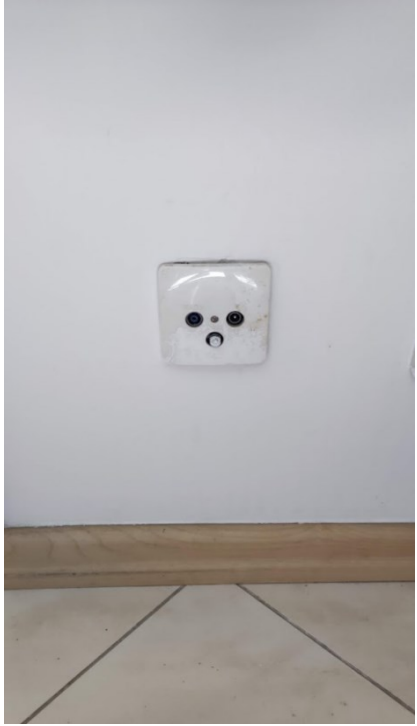
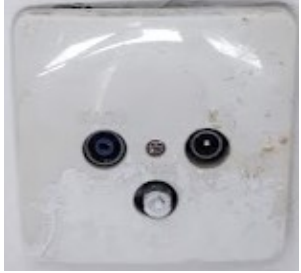
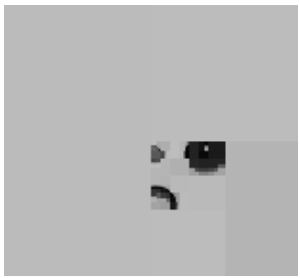
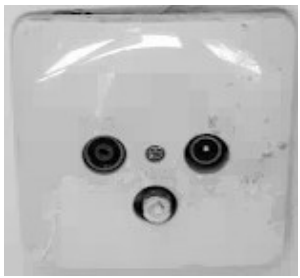
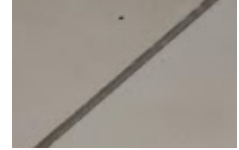
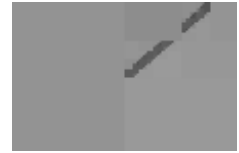
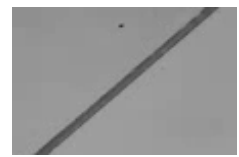

t3)

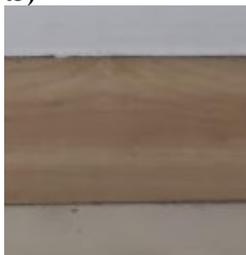
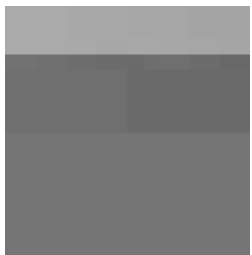
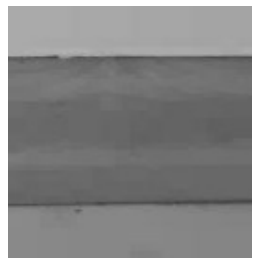

Figure 1: Socket source image and templates

**Source image**        t1)        t2)

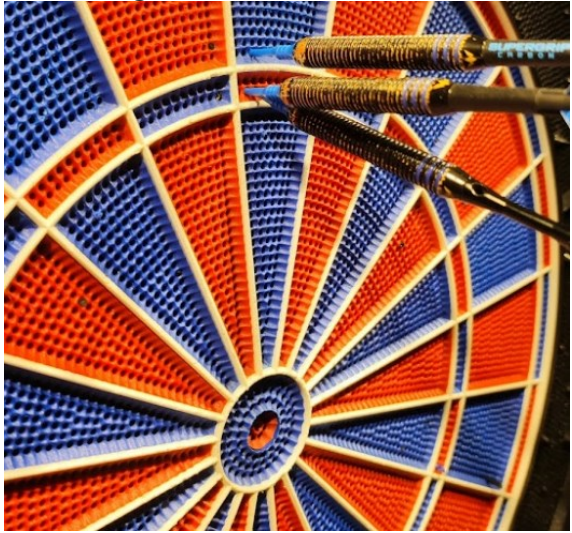
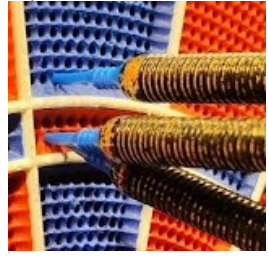
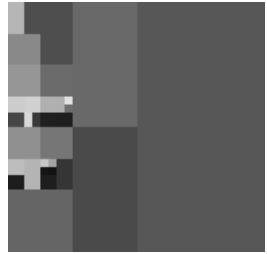
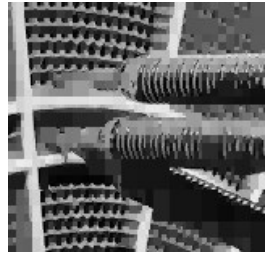
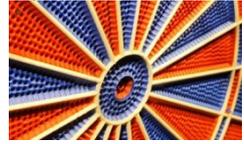
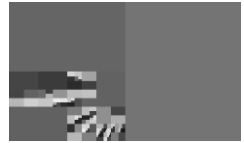
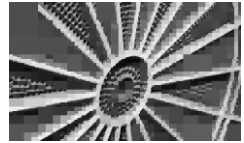

t3)

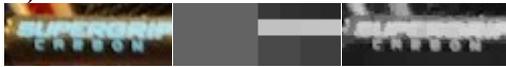

Figure 2: Darts source image and templates

**Source image**

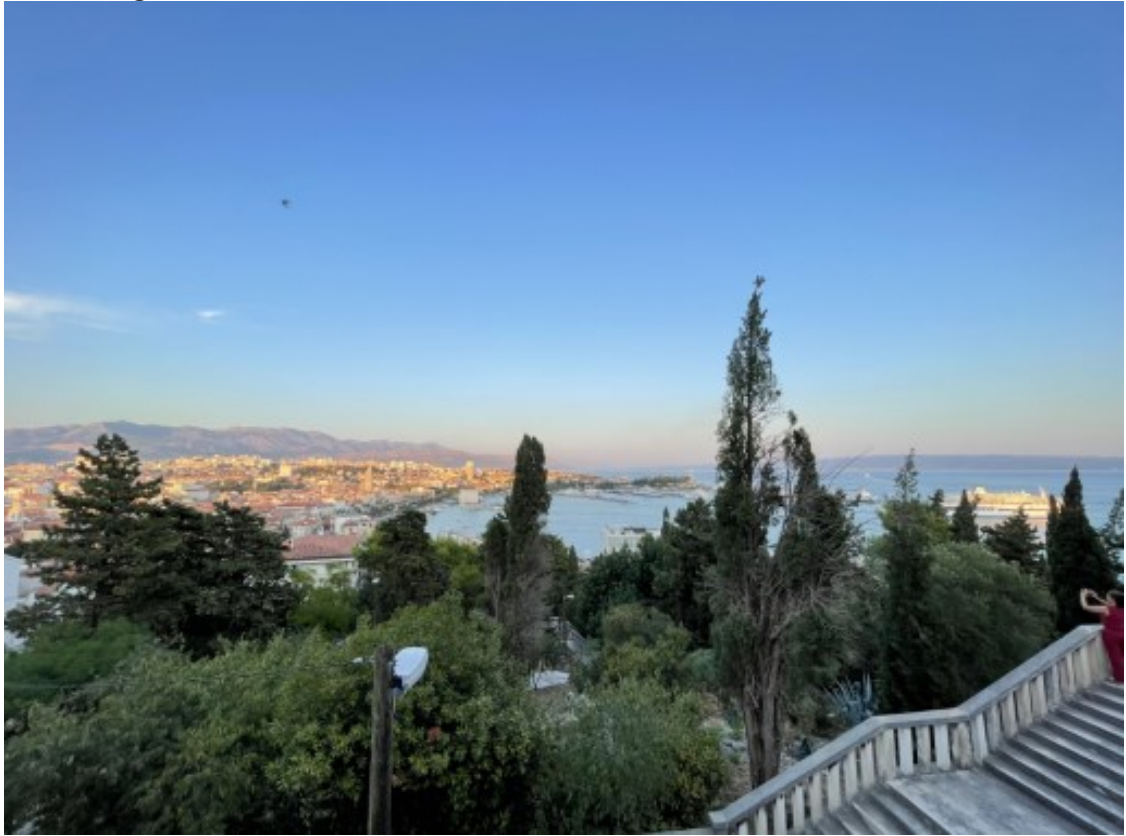

t1)

t2)
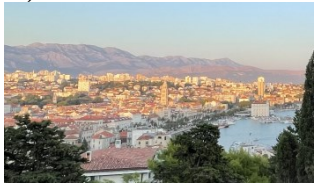
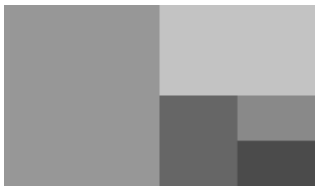
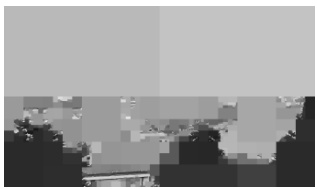

t3)
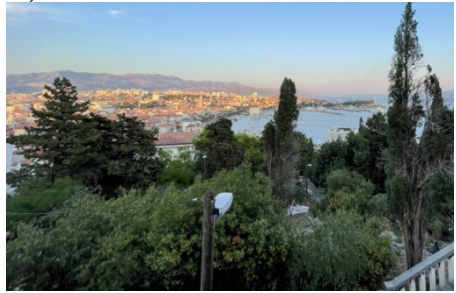
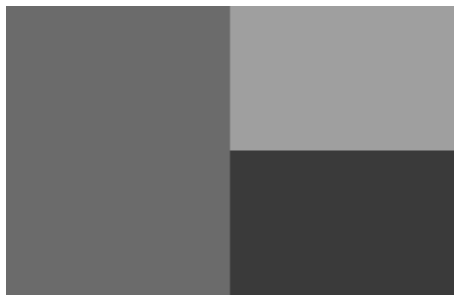
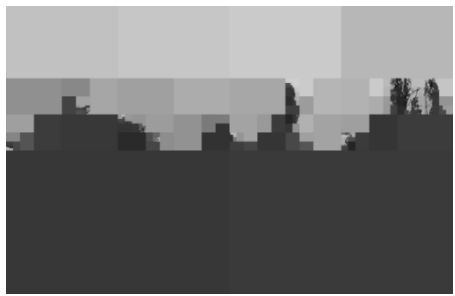

t4)
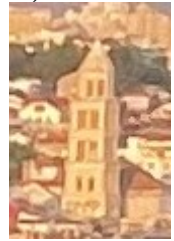
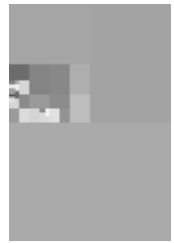
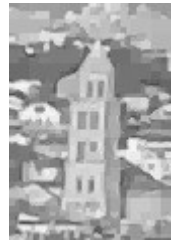

**t5)**

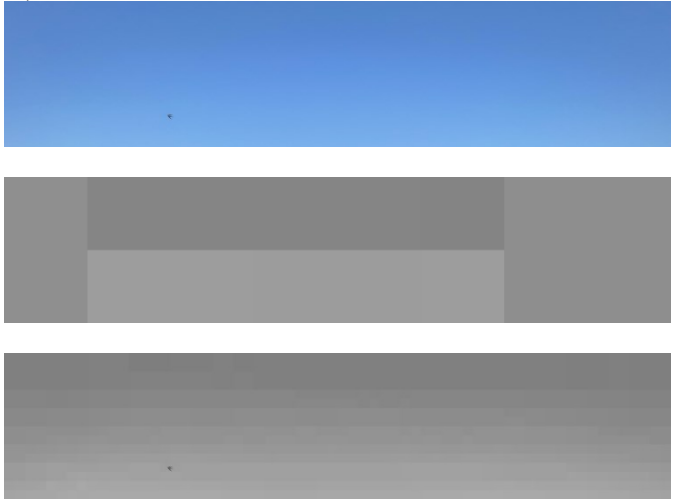

Figure 3: Split city source image and templates